\pdfoutput=1
\documentclass[11pt]{article}

\usepackage[preprint]{acl}

\usepackage{times}
\usepackage{latexsym}

\usepackage[T1]{fontenc}
\usepackage[utf8]{inputenc}

\usepackage{microtype}

\usepackage{inconsolata}

\usepackage{graphicx, url, amsmath, multirow, multicol, array, float, subfig, comment}

\title{Grammatical Error Feedback: An Implicit Evaluation Approach}

 \author{Stefano Bannò, Kate Knill, Mark J. F. Gales \\
         ALTA Institute, Department of Engineering, University of Cambridge, UK \\ \texttt{\{sb2549,kmk1001,mjfg100\}@cam.ac.uk}}

\begin{document}
\maketitle
\begin{abstract}
Grammatical feedback is crucial for consolidating second language (L2) learning. Most research in computer-assisted language learning has focused on feedback through grammatical error correction (GEC) systems, rather than examining more holistic feedback that may be more useful for learners. This holistic feedback will be referred to as grammatical error feedback (GEF). In this paper, we present a novel implicit evaluation approach to GEF that eliminates the need for manual feedback annotations. Our method adopts a grammatical lineup approach where the task is to pair feedback and essay representations from a set of possible alternatives. This matching process can be performed by appropriately prompting a large language model (LLM). An important aspect of this process, explored here, is the form of the lineup, i.e., the selection of foils. This paper exploits this framework to examine the quality and need for GEC to generate feedback, as well as the system used to generate feedback, using essays from the Cambridge Learner Corpus.

%% of correct and incorrect essays, generating multiple learner essay versions with varying grammatical error rates. We prompt a large language model (LLM) to produce natural-language feedback based on the differences between these versions and their corrected counterparts. Finally, we use another LLM to evaluate the consistency between essay versions and generated feedback using two evaluation methods.

\end{abstract}

\section{Introduction}

In natural language processing, computer-assisted language learning (CALL) is a well-established research area~\cite{chapelle2001computer}. An important part of this process is to give feedback on the use of grammar by a learner. This feedback has usually been in the form of grammatical error detection (GED)~\cite{leacock2014automated}, and grammatical error correction (GEC)~\cite{wang2021comprehensive, bryant2023}, and the latter has been the subject of four shared tasks over the past 15 years. While highlighting errors and providing a grammatically corrected text is beneficial for second language (L2) learners~\cite{hyland2006}, it is even more valuable to offer specific feedback that comments on grammatical errors, explains them, and provides suggestions for improvement. As shown in \citet{ellis2006implicit}, giving metalinguistic explanations (i.e., \emph{explicit} feedback) helps learners understand their errors and learn how to avoid them in the future in a more effective way than providing mere recasts (i.e., \emph{implicit} feedback).

The recent advent of large language models (LLMs) has significantly altered the world of CALL. LLMs have been investigated for L2 holistic~\cite{yancey2023rating} and analytic~\cite{banno-etal-2024-gpt} assessment as well as GEC~\cite{coyne2023analyzing, fang2023chatgpthighlyfluentgrammatical, wu2023chatgptgrammarlyevaluatingchatgpt, katinskaia-yangarber-2024-gpt}. An interesting aspect of natural language generation (NLG) models, such as LLMs, is that they offer the opportunity to automatically generate free-form feedback for candidates, which will be referred to as grammatical error feedback (GEF) hereafter. GEF differs from GED and GEC in that it is not meant to provide feedback on all errors made, but rather provide a more holistic and compact interpretation of the errors being made, along with explanations of the error and suggested improvements.  

One of the challenges for developing GEF systems is that assessment is difficult. For GED and GEC, it is possible to generate reference transcriptions against which systems can be assessed. Given the vast array of appropriate grammatical feedback that can be generated for an essay, it is not feasible to generate a set of reference feedback that sufficiently covers the complete spectrum of possible valid output. 
To address this problem in this work, we introduce a novel implicit evaluation approach to GEF that does not require manual annotations. Rather than explicitly evaluating the feedback, the proposed scheme adopts a {\it grammatical lineup} approach similar to voice lineups in forensic speaker recognition~\cite{mcdougall2013assessing}. The task is then to appropriately pair feedback and essay from the possible set of comparisons.

An important aspect of any lineup is the generation of the foils, i.e., the alternatives to the matched pair. There are a number of considerations for the selection of a foil. It is important that the attribute being assessed is distributed over the lineup. For feedback it is thus important to avoid, or at least assess the effect of, differing content as the system may have a bias to particular content, such as performing simple lexicon matching in pairs rather than assessing feedback. Another crucial stage in this process is the selection stage, obtaining the best matching pair. Again, this can exploit the emergent properties of LLMs.

In the next section, we briefly review related literature. Section \ref{sec:grammatical_error_feedback} explains the concept of GEF in more detail and Section~\ref{sec:lineup} presents grammatical lineups.
In Section \ref{sec:data}, we describe the data used in our experiments and, in Section \ref{sec: experimental}, we go through each step of our proposed pipeline. Experimental results are presented in Section \ref{sec:results} and discussed in Section \ref{sec:discussion}. Finally, in Section \ref{sec:conclusions}, we move on to the conclusions and discuss potential future directions.

\section{Related work}
\label{sec:related}

For grammatical error annotation in CALL, the ERRor ANnotation Toolkit (ERRANT)~\cite{bryant2017automatic} has become a widely-used tool for extracting and categorising grammatical error information. Specifically, it extracts error edits from parallel original and corrected sentences. Examples of ERRANT edit labels are \texttt{R:VERB:FORM}, which indicates the need to replace an incorrect verb form, and \texttt{M:DET}, which indicates a missing determiner. The prefix \texttt{R:} stands for \emph{replace}, \texttt{M:} for \emph{missing}, and \texttt{U:} for \emph{unnecessary} (see Section \ref{sec:lineup} for an example). ERRANT labels error types as \texttt{OTHER} when edits do not fall under any other category. A large part of errors labelled as \texttt{OTHER} are paraphrases. ERRANT has become a standard approach for GEC, but its output information, included in a so-called \( M^2 \) file, is not always easily readable, and, despite categorising errors based on parts of speech, it does not provide clear natural-language-based descriptions of errors or motivate corrections in a comprehensible way for learners and teachers.

Other previous works on feedback generation followed the organisation of a shared task on feedback comment generation for language learning called ~\emph{FCG GenChal 2022}~\cite{nagata-etal-2021-shared, nagata-etal-2023-report}. The main limitations of this task were that it only targeted preposition errors and only leveraged incorrect sentences tagged with span-based error annotations as inputs, hence without their respective corrections. Participants such as \citet{behzad-etal-2023-sentence}, \citet{stahl-wachsmuth-2023-identifying}, \citet{jimichi-etal-2023-feedback}, \citet{ueda-komachi-2023-tmu}, and \citet{koyama-etal-2023-tokyo} tackled this problem by fine-tuning systems based on T5~\cite{raffel2020exploring}, BART~\cite{lewis-etal-2020-bart}, and GPT-2~\cite{radford2019gpt2}.  In the context of this shared task, \citet{coyne-2023-template} and \citet{coyne2023developing} developed a bipartite typology for L2 feedback which includes abstract tags (e.g., idiom, language transfer, praise, etc.) and grammatical pattern tags (e.g., conditional, possessive, relative clause, etc.).

%% The recent advent of large language models (LLMs) has significantly altered the world of CALL. LLMs have been investigated for L2 holistic~\cite{yancey2023rating} and analytic~\cite{banno-etal-2024-gpt} assessment as well as GEC~\cite{coyne2023analyzing, katinskaia-yangarber-2024-gpt}.
A recent study explored the application of LLMs for grammatical error explanation~\cite{song-etal-2024-gee}. The study employed a two-step pipeline that first used fine-tuned and prompted LLMs to extract structured atomic token edits. Subsequently, GPT-4~\cite{openai2024gpt4technicalreport} was prompted to explain each edit. Despite its novelty and appeal, this approach has several limitations, i.e., it a) can only be applied to sentences but not longer compositions; b) only consists of four operation edit-level types, namely \emph{insert}, \emph{delete}, \emph{replace}, and \emph{relocate}; and c) needs manual annotators to check that the error explanations are correct, which is generally a costly operation. It is also worth mentioning the recent work by \citet{stahl-etal-2024-exploring} on joint essay scoring and feedback generation, although it focuses on L1 essays.

%% In this work, we introduce a novel implicit evaluation approach to grammatical error feedback (GEF) in a compact, comprehensive, and free-form manner that does not require manual annotations on feedback responses. Instead, our method relies on pairs of parallel correct and incorrect essays. We create multiple versions of learner essays with varying grammatical error rates through editing. After augmenting the dataset in this manner, in addition to manual GEC annotations, we also use automatic GEC systems to correct the essays. Next, we prompt a large language model (LLM) to generate natural-language feedback based on the differences between the learner essays and their corrected versions. Finally, we utilise another LLM to assess the alignment between the various essay versions, their corresponding ERRANT information, and the generated feedback responses. To do this, we employ two different evaluation methods. One is based on the various essay versions, whereas the other one revolves around feedback responses derived from the various essay versions.

%% In Section \ref{sec:data}, we describe the data used in our experiments. In Section \ref{sec: experimental}, we go through each step of our proposed pipeline, while we illustrate the experimental results in Section \ref{sec:results}, which are then discussed in Section \ref{sec:discussion}. Finally, in Section \ref{sec:conclusions}, we move on to the conclusions and discuss potential future directions.

\section{Grammatical error feedback}
\label{sec:grammatical_error_feedback}
The aim of grammatical error feedback (GEF) is to supply holistic feedback to the user on their use of grammar.
This feedback should summarise the forms of grammatical error being made in easy to interpret, informative, form, not simply listing all the errors being made by the candidate in their essay (see Appendix \ref{sec:appendix_b} for an example). As the form of feedback will be free-form in nature, NLG conditioned on the essay is used with the associated flexibility of response. The standard approach to this, and the one adopted in this work, is the use of LLMs with the appropriate prompt and inputs. 

\begin{figure}[!htbp]
    \centering
    \includegraphics[width=0.85\linewidth]{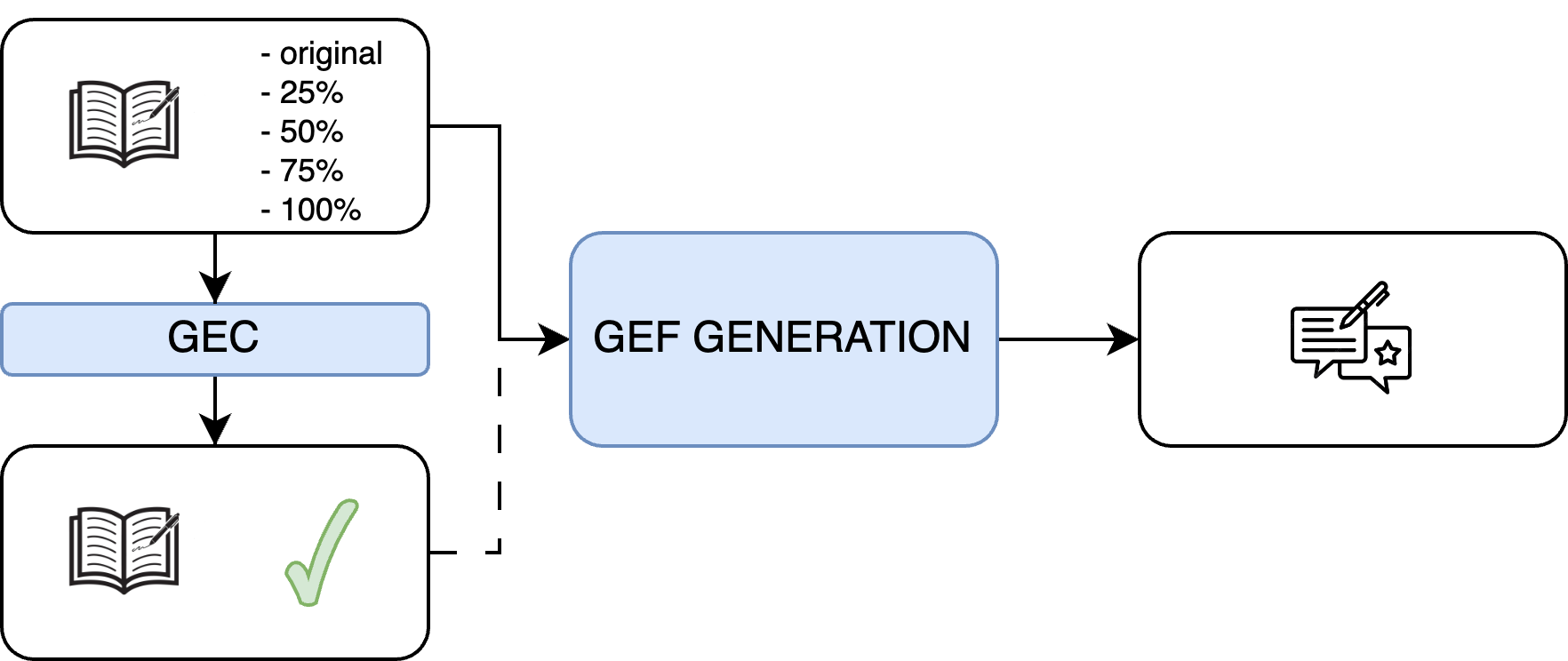}
    \caption{GEF generation, with/without explicit GEC. 
 %%    (marked by the tick).
    }
    \label{fig:gef_gen_step}
\end{figure}

Since LLMs, especially the very large LLMs currently available via APIs, are capable of performing GEC~\cite{katinskaia-yangarber-2024-gpt}, they should be able to detect the grammatical errors as part of the feedback generation process.
The alternative, related to chain-of-thought processing~\cite{wei2022chain}, is to specify that the model, or an alternative system, performs GEC, and the corrected essay is explicitly used for the GEF generation process (see Figure \ref{fig:gef_gen_step}). 

\begin{comment}
\begin{figure}[!htbp]
    \centering
    \includegraphics[width=1.0\linewidth]{gec.png}
    \caption{GEC.}
    \label{fig:gec_step}
\end{figure}
\end{comment}

The focus in this work is not the prompt design per se but rather the information that is required to give good feedback, particularly whether it is necessary to perform GEC and feed both the original learner essay and the GEC version into the GEF generation process. Thus, reasonable forms of prompt were adopted (see  
Appendix \ref{sec:appendix_a}) and tuned to yield feedback in an easily interpretable summative form, rather than attempting to optimising the form of prompt\footnote{Though the assessment approaches described in the next section can be used to refine the prompts.}.
%In addition to the question of whether explicit GEC is required, 
It is also interesting to consider the quality of the GEC system and how it impacts the feedback. Note examining the impact of GEC in this fashion is not the same as using standard GEC metrics, as it only evaluates the influence on feedback generation. This difference means that some of the issues with standard LLM-based GEC systems~\cite{fang2023chatgpthighlyfluentgrammatical, wu2023chatgptgrammarlyevaluatingchatgpt} may not occur.

\section{Grammatical lineup}
\label{sec:lineup}
\begin{figure*}[!htbp]
    \centering
    \includegraphics[width=0.525\linewidth]{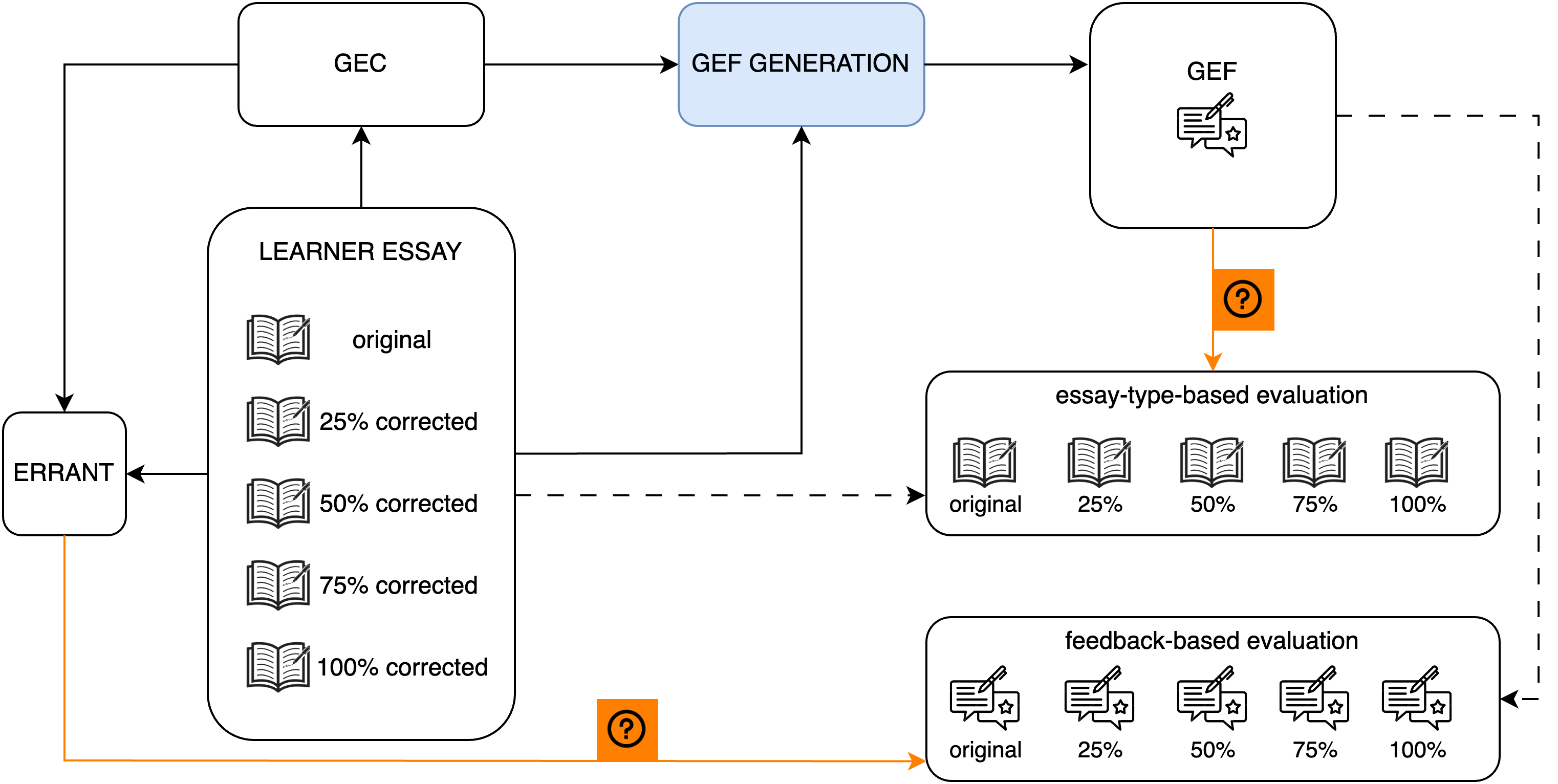}
    \caption{Proposed framework.}
    \label{fig:pipeline}
\end{figure*}

As previously discussed, generating a set of references of free-form feedback for an essay is highly challenging given the wide-range of possible forms of feedback possible. Rather than adopting an explicit error measure, as done in GED and GEC, for GEF an implicit performance metric based on matching against a grammatical lineup is proposed. This requires no manual references, but does require the generation of a suitable lineup, i.e., a set of foils, that enables the assessment of GEF quality. The overall framework is shown in Figure~\ref{fig:pipeline}.

The generation of a suitable set of foils is the most important choice in the lineup. For this work to ensure that there are no biases introduced by the essay content, sets of essays are generated based on different levels of manual correction of the learner essay. Here, five different versions of the essay were used:\footnote{This choice of lineup generation assumes that there is a range of numbers of grammatical errors in the essays from the corpus. It removes any biases that may result from artificially adding grammatical errors to an essay, but does limit the range of assessment. The random correction, in comparison to accounting for the ability of candidates to address a grammatical structure, is another limitation, but is not expected to bias the evaluation process.}
\begin{itemize}
    \item \emph{original}: the original essay as written by the learner;
    \item \emph{25\% corrected}: a version where only 25\% of the errors are randomly corrected;
    \item \emph{50\% corrected}: a version where only 50\% of the errors are randomly corrected;
    \item \emph{75\% corrected}: a version where only 75\% of the errors are randomly corrected;
    \item \emph{100\% corrected}: a version where all errors are corrected.
\end{itemize}
It is then possible to use this set of essays as the basis for both the generation of the parade, i.e. the foils, as well as the feedback. The task will then be to match both the feedback and parade entry from the same essay. The quality of the matching process, indicated by the "{\tt ?}" blocks in Figure~\ref{fig:pipeline}, is important for the assessment process. Rather than asking the system to select over a range, it uses a simple binary selection process of asking whether the feedback and essay representation match. This is motivated by the use of LLMs for comparative rather than absolute assessment~\cite{liusie-etal-2024-llm}.

Another important aspect of designing the parade is to ensure that the quality of the feedback is being assessed rather than simply matching lexical entries in the parade and the feedback. Good feedback is expected to make use of examples of grammatical errors in the essay. This can easily bias the process as the essays and feedback do not contain the same grammatical errors by design. Two forms of parade are investigated in this work. 

\begin{figure}[!htbp]
    \centering
    \includegraphics[width=1.0\linewidth]{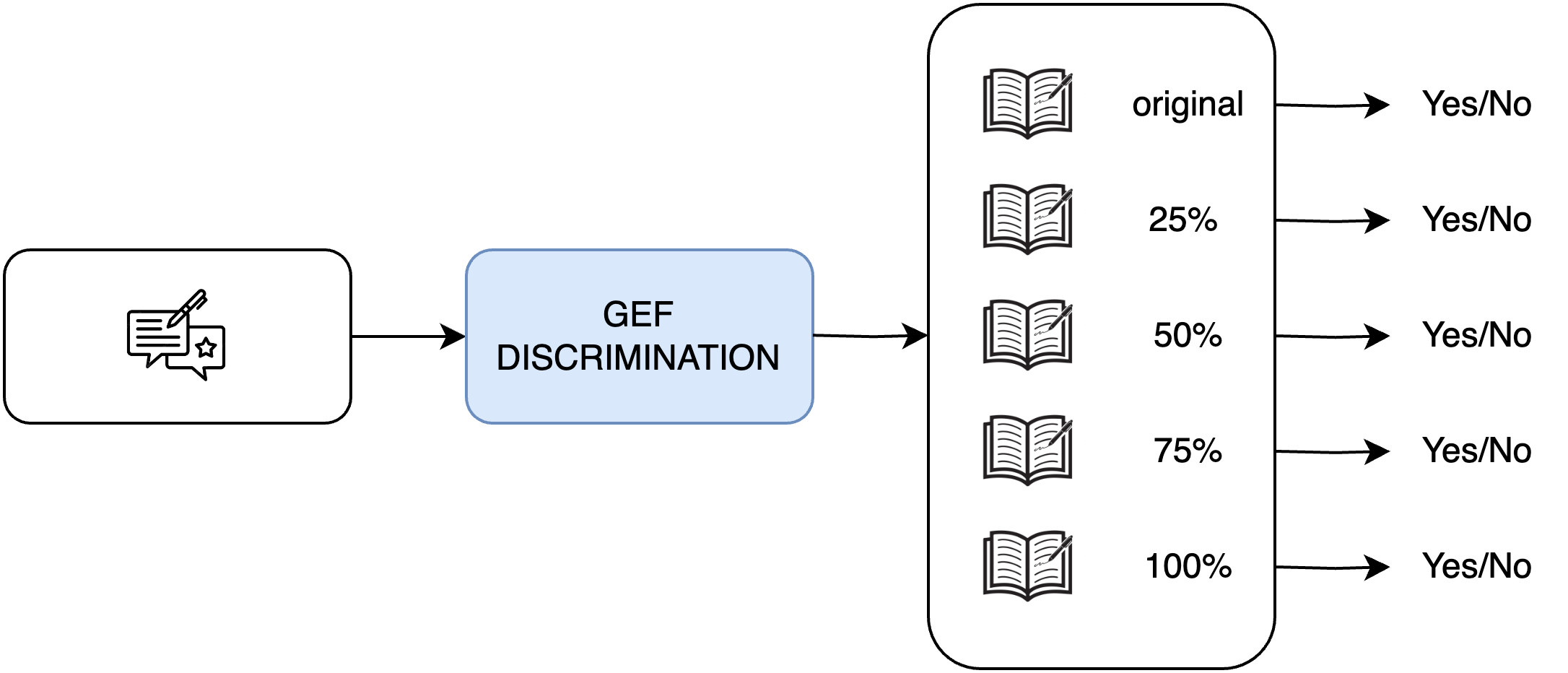}
    \caption{GEF: essay-type-based evaluation.}
    \label{fig:gef_discr_essay}
\end{figure}

\paragraph{Essay-type-based evaluation:} In this configuration, we ask the LLM to decide whether a given feedback response matches a given essay using a lineup that includes all essay versions.
%% i.e., \emph{original}, \emph{25\% corrected}, \emph{50\% corrected}, \emph{75\% corrected}, and \emph{100\% corrected}.
Each essay version is fed into the system one at a time, and the LLM should simply output ``yes'' or ``no''. For each feedback-essay pair, we extract the probability of ``yes'' being the first predicted token. The prompt can be found in Appendix \ref{sec:appendix_a}, and this type of evaluation is illustrated in Figure \ref{fig:gef_discr_essay}.

This approach ensures that all the information that was available to generate the feedback is available for matching. However, as discussed, there is the option of the matching process exploiting lexical matches in the feedback and the essay, rather than specifically assessing whether the holistic feedback matches.

\begin{figure}[!htbp]
    \centering
    \includegraphics[width=1.0\linewidth]{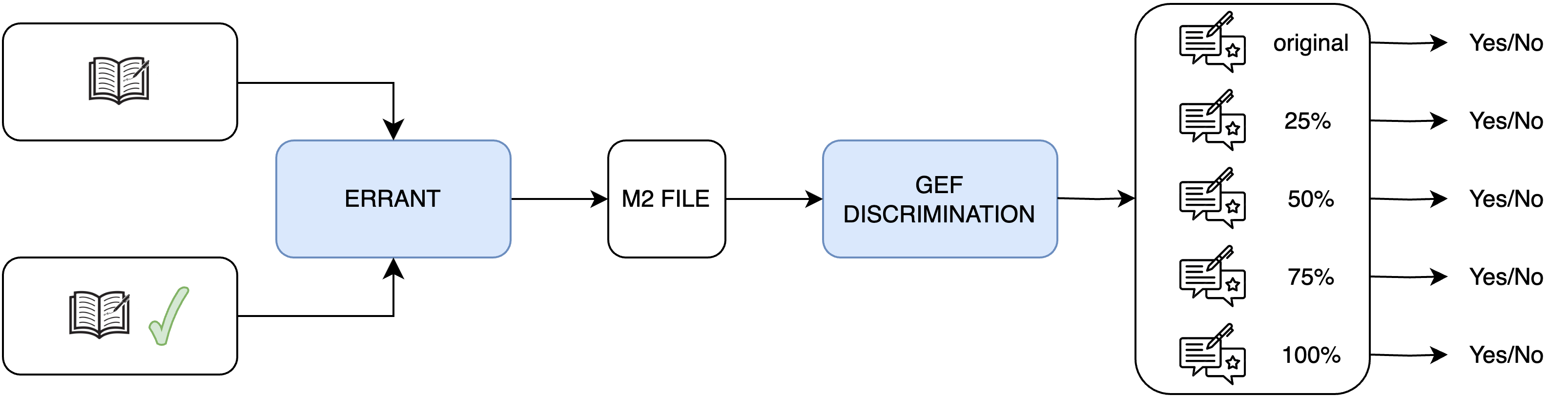}
    \caption{GEF: feedback-based evaluation.}
    \label{fig:gef_discr_feedback}
\end{figure}

\paragraph{Feedback-based evaluation:} Conversely, in this configuration, we ask the LLM to decide whether a given ERRANT \( M^2 \) file matches a given feedback response using a lineup that contains all feedback responses from the set of essays.
%% i.e., the ones obtained from \emph{original}, \emph{25\% corrected}, \emph{50\% corrected}, \emph{75\% corrected}, and \emph{100\% corrected}.
Using only the grammatical errors restricts the information used in matching to only grammatical information rather than any other information in the essay. This matching process is shown in Figure~\ref{fig:gef_discr_feedback}.

However, using the original form of the \( M^2 \) file does not address possible lexical matching, as shown in the following snippet.

{\tiny
\begin{verbatim}
     S Hello Mike , I bought a mobile phone , I like it because I can liste to  
    music and I can see funny videos . It is expensive but it is good and
    it is black . Write soon .
    A 15 16 |||R:SPELL|||listen|||REQUIRED|||-NONE-|||0
    A 21 22 |||R:VERB|||watch|||REQUIRED|||-NONE-|||0
    A 26 27 |||R:VERB:TENSE|||was|||REQUIRED|||-NONE-|||0
\end{verbatim}
}
%% As can be seen in Figure \ref{fig:m2_file_example}, original \( M^2 \) files also include the original learner text in the first line, but we removed it in order not to lexically bias the LLM.

\noindent
In contrast to the essay-type based assessment, it is possible to modify or remove the lexicon entries in the \( M^2 \) file. Here, three forms are examined:

     a) {\it standard:} the standard \( M^2 \) file;

    b) {\it  replaced corrected word:} an \( M^2 \) file where corrected words are replaced with the respective incorrect words from the learner essay;

    c) {\it no lexical information:} an \( M^2 \) file where corrected words are removed.\footnote{Note that the first line containing the learner essay is removed from all three formats. Examples can be found in Appendix \ref{sec:appendix_c}.}

\noindent
Comparing the performance of these options allows an analysis of the level of lexical information being exploited by the matching process. Similarly to the essay-type-based evaluation, for each \( M^2 \) file-feedback pair, we extract the probability of ``yes'' being the first predicted token. The prompt can be found in Appendix \ref{sec:appendix_a}.\footnote{Note that both evaluation methods are reversible, i.e., the former can revolve around GEF responses instead of essays, and the latter around essay representations (i.e, \( M^2 \) files) instead of GEF responses. We observed the same trends also when reversing them, but we decided to only report the results related to the current configurations due to space constraints.}

\section{Data}
\label{sec:data}
The Cambridge Learner Corpus (CLC) is an ever-growing collection of L2 written data obtained from Cambridge English exams. According to \citet{okeefe2017english}, as of 2017, it included 266,600 examination scripts and 143 L1 backgrounds collected in the period between 1993 and 2012. The corpus includes data from lower to advanced proficiency levels annotated according to the Common European Framework of Reference (CEFR)~\citep{cefr2001} and features manual annotations with information about errors according to a taxonomy of about 80 error types described in \citet{nicholls2003cambridge}. An example drawn from the data in XML format is the following:
\begin{itemize}
    \item [] \emph{I am looking forward to} $<$NS type=``FV''$>$ \emph{hear $|$ hearing} $<$/NS$>$ \emph{from you.}
\end{itemize}
in which FV indicates a verb form error and \emph{hear} is corrected to \emph{hearing}.

For our experiments, we extracted 300 essays written by learners representing 28 L1 backgrounds and 40 nationalities. To ensure representative data across proficiency levels, we selected 50 essays per proficiency level ranging from A1 to C2.
%We chose this dataset instead of the publicly available CLC-FCE~\citep{yannakoudakis2011}, as the latter exclusively includes learners at the B2 level.
%To enable different levels of grammatical error required for the lineup, the XML structure of the data is ideal because it allows us to manipulate the quantity of errors and corrections in an easy and flexible manner.

\section{Experimental setup}
\label{sec: experimental}

As should be clear at this point, we can identify three major steps in our proposed pipeline: GEC, GEF generation, and GEF discrimination.

\paragraph{GEC:} In addition to manual corrections, we use automatic corrections obtained employing GECToR~\cite{omelianchuk-etal-2020-gector}, Gramformer\footnote{\url{https://github.com/PrithivirajDamodaran/Gramformer}} -- which is a publicly available T5-based GEC system -- and GPT-4o. In order to have an exhaustive comparison, each system should represent a different category of GEC systems according to the taxonomy illustrated in \citet{omelianchuk-etal-2024-pillars}, i.e., edit-based systems, sequence-to-sequence models, and LLM-based systems, respectively. Each of the 5 essay versions -- including the \emph{100\% corrected} one -- is fed into the GEC systems to obtain the respective grammatically corrected version.

Both GECToR -- with RoBERTa~\cite{liu2019robertarobustlyoptimizedbert} as pretrained encoder -- and Gramformer are used off-the-shelf. The essays are segmented into sentences before being fed into these models to exploit them under optimal conditions, as their maximum sequence lengths are 50 and 64 tokens, respectively. To segment the essays, we use the spaCy\footnote{\url{https://spacy.io/}} sentencizer and then double-check and fix the obtained sentences manually. Conversely, when we use GPT-4o, we directly feed the entire essays into the model. The prompt can be found in Appendix \ref{sec:appendix_a}.

To evaluate GEC, we use the ERRANT F$_{0.5}$ score as the primary metric, which is calculated by extracting the GEC edits and then comparing the reference and hypothesis. Additionally, we use the Generalized Language Evaluation Understanding (GLEU) metric~\cite{napoles2015ground}, which is inspired by BLEU~\cite{papineni2002bleu} and rewards the overlap of \emph{n}-grams between the correction and the reference, while penalising \emph{n}-grams in the correction that remain unchanged when they have been altered in the reference.

\paragraph{GEF generation:} Each version of the essay and its respective corrected version are fed into an LLM in order to obtain a natural language feedback response. We consider three LLMs for our experiments, i.e., Llama 3 8B~\cite{llama3}, GPT-3.5 (\emph{gpt-3.5-turbo-0125})~\cite{brown2020language}, and GPT-4o. In order to check whether the information derived from GEC is useful, we also ask the models to generate feedback only based on the essays without input of their respective corrected versions. The prompts used for GEF generation can be found in Appendix \ref{sec:appendix_a}.

\paragraph{GEF discrimination:} For each essay, we now have a natural language GEF response. An example can be found in Appendix \ref{sec:appendix_b}. To check for feedback correctness, we proceed to the next step, which we call GEF discrimination, implemented using another LLM with the two different evaluation methods described in Section \ref{sec:lineup}, i.e., essay-type-based evaluation and feedback-based evaluation. As in the previous step, we consider Llama 3 8B, GPT-3.5, and GPT-4o.

\smallskip

The evaluation metric employed for GEF discrimination is Accuracy. Specifically, for the essay-type-based evaluation method, let:

\[
E = \{ E_0, E_{25}, E_{50}, E_{75}, E_{100} \} \quad
\]

be the set of essay versions (i.e., \emph{original}, \emph{25\% corrected}, \emph{50\% corrected}, \emph{75\% corrected}, and \emph{100\% corrected}) and $F$ a given GEF response.

For each essay version \( E_i \), we calculate the probability:

\[
P(E_i | F) \quad \text{ for } i = 0, 25, 50, 75, 100
\]

and then we compute the Accuracy $Acc$:

\[
Acc = \frac{C}{N} \quad \text{with } \hat{E} = \arg\max_{i} P(E_i | F)
\]

where \( C \) is the count of correctly predicted matches and \( N \) is the total number of GEF responses tested.

Similarly, for the feedback-based evaluation method, let:

\[
F = \{ F_0, F_{25}, F_{50}, F_{75}, F_{100} \} \quad
\]

be the set of GEF responses obtained from the set of essay versions and \( M^2 \) the ERRANT \( M^2 \) file associated to a given essay.

For each GEF response \( F_i \), we calculate the probability:

\[
P(F_i | M^2) \quad \text{ for } i = 0, 25, 50, 75, 100
\]

and then we compute the Accuracy $Acc$:

\[
Acc = \frac{C}{N} \quad \text{with } \hat{F} = \arg\max_{i} P(F_i | M^2)
\]

where \( C \) is the count of correctly predicted matches and \( N \) is the total number of \( M^2 \) files, hence essays, tested.

For both GEF discrimination methods, we initially tried a comparative framework in which we fed all essay versions or feedback responses at once and asked the LLM to select the most suitable option, but it was found to be too difficult a task.

\section{Experimental results}
\label{sec:results}

\paragraph{GEC results:} While GEC is not the main focus of this work, it is important to briefly discuss these results, as they will partly influence the outcomes obtained in the subsequent steps.

Table \ref{tab:gec_results} reports the ERRANT F$_{0.5}$ scores when comparing the automatic corrections from our three GEC systems across all the essay versions (except for \emph{100\% corrected}) to the manual references. The results show that GPT-4o achieves the best results on the \emph{original} and \emph{25\% corrected} essays, whereas GECToR performs better on the \emph{50\% corrected} and \emph{75\% corrected} essays. This discrepancy is most likely due to GPT’s tendency to overcorrect grammatical errors~\cite{fang2023chatgpthighlyfluentgrammatical, wu2023chatgptgrammarlyevaluatingchatgpt} as it appears from the high Recall scores (see Tables \ref{tab:gec_res_full_manual}, \ref{tab:gec_res_full_gpt4o}, \ref{tab:gec_res_full_gector}, and \ref{tab:gec_res_full_gram} in Appendix \ref{sec:appendix_d}).

\begin{table}[h!]
  \centering
  \small
  \textbf{GPT-4o}
  \begin{tabular}{c|c|c|c|c}
    \hline
    \multirow{2}{*}{\textbf{HYP/REF}} & \multicolumn{4}{c}{\textbf{Manual}} \\ \cline{2-5}
                                      & \textbf{Original} & \textbf{25\%} & \textbf{50\%} & \textbf{75\%} \\ \hline
    \textbf{Original} & \textbf{43.55} & 18.00 & 8.57 & 4.69 \\ \hline
    \textbf{25\%} & 18.86 & \textbf{40.28} & 8.42 & 4.59 \\ \hline
    \textbf{50\%} & 10.22 & 9.32 & 33.67 & 4.79  \\ \hline
    \textbf{75\%} & 6.20 & 5.51 & 5.19 & 23.14  \\
    
  \end{tabular}
\newline
\vspace*{0.2 cm}
\newline
\textbf{GECToR}
\begin{tabular}{c|c|c|c|c}
    \hline
    \multirow{2}{*}{\textbf{HYP/REF}} & \multicolumn{4}{c}{\textbf{Manual}} \\ \cline{2-5}
                                      & \textbf{Original} & \textbf{25\%} & \textbf{50\%} & \textbf{75\%}\\ \hline
    \textbf{Original}                 & 40.90 & 17.51 & 8.16 & 4.59 \\ \hline
    \textbf{25\%}                     & 18.03 & 39.22 & 7.56 & 4.92  \\ \hline
    \textbf{50\%}                     & 9.54 & 8.77 & \textbf{33.90} & 4.67 \\ \hline
    \textbf{75\%}                     & 5.80 & 5.59 & 4.77 & \textbf{26.10} \\ 
  \end{tabular}
\newline
\vspace*{0.2 cm}
\newline
\textbf{Gramformer}

\begin{tabular}{c|c|c|c|c}
    \hline
    \multirow{2}{*}{\textbf{HYP/REF}} & \multicolumn{4}{c}{\textbf{Manual}} \\ \cline{2-5}
                                      & \textbf{Original} & \textbf{25\%} & \textbf{50\%} & \textbf{75\%} \\ \hline
    \textbf{Original}                 & 35.75 & 15.39 & 8.00 & 4.63 \\ \hline
    \textbf{25\%}                     & 15.83 & 34.27 & 7.19 & 4.50  \\ \hline
    \textbf{50\%}                     & 8.86 & 8.22 & 31.31 & 4.32 \\ \hline
    \textbf{75\%}                     & 5.52 & 5.44 & 4.76 & 22.65 \\ 
  \end{tabular}
  \caption{ERRANT F$_{0.5}$ scores for Auto GEC (GPT-4o, GECToR, and Gramformer) vs Manual.}
  \label{tab:gec_results}
\end{table}

These results are confirmed when we evaluate the performance of the three systems in terms of GLEU, as can be seen in Table \ref{tab:res_gleu}, where we compare the source and hypothesis of each essay version to the corrected manual reference. In this case, even Gramformer outperforms GPT-4o on the \emph{50\% corrected} and \emph{75\% corrected} essay versions.

\begin{table}[h!]
  \centering
  \footnotesize
  \begin{tabular}{c||c||c|c|c}
  
    & \textbf{lower} & \multirow{2}{*}{\textbf{GPT-4o}} & \multirow{2}{*}{\textbf{GECToR}} & \multirow{2}{*}{\textbf{Gramformer}} \\
    & \textbf{bound} & & & \\ \hline\hline
    \textbf{Orig.} & 66.30 & \textbf{78.26} & 75.97 & 73.53 \\ \hline
    \textbf{25\%} & 73.02 & \textbf{80.95} & 80.17 & 78.18 \\ \hline
    \textbf{50\%} & 81.66 &  83.58 & \textbf{85.11} & 84.25 \\ \hline
    \textbf{75\%} & 89.83 &  86.07 & \textbf{89.92} & 89.45 \\
     
  \end{tabular}
  \caption{GLEU scores for various essay versions for Auto GEC on GPT-4o, GECToR, and Gramformer.}
  \label{tab:res_gleu}
\end{table}

\paragraph{GEF results:} We now move on to the core part of this section, where we report and analyse the GEF results, starting from a comparison of the impact of all GEC systems when using GPT-4o both for GEF generation and GEF discrimination (see Table \ref{tab:gef_res_essay_gpt4}). For the essay-type-based evaluation method, the results show that the best performance is achieved when using GECToR as the GEC system. This indicates that potential self-bias in the GEF generation phase is ruled out, as GECToR even outperforms GPT-4o. However, as mentioned in Section \ref{sec: experimental}, the reader should take into account the high presence of lexical information since the GEF response is matched directly to the essay versions when employing this evaluation method. Therefore, under these circumstances, even the No GEC system achieves acceptable results.

\begin{table}[h!]
\small
\centering
\begin{tabular}{c||c}

%& \textbf{essay-type-based}  \\
%\hline
\textbf{Form of GEC} & \textbf{\% Accuracy} \\
\hline\hline
         Manual & 61.26         \\
\hline 
         GPT-4o & 56.66              \\
%\hline
         GECToR &  \textbf{59.93}            \\
%\hline
         Gramformer &  57.26          \\
\hline
         No GEC &  57.46         \\

\end{tabular}
\caption{Overall Accuracy (\%) results for \textbf{essay-type-based} evaluation when using GPT-4o both for GEF generation and discrimination.}
\label{tab:gef_res_essay_gpt4}
\end{table}

For this reason, we also consider the feedback-based evaluation method, for which the results are reported in Table \ref{tab:gef_res_feedback_gpt4}.
\begin{table}[h!]
\centering
\small
\begin{tabular}{c||c|c|c}

& \multicolumn{3}{c}{\textbf{\% Accuracy}} \\
\textbf{Form of GEC}& \textbf{standard} & \textbf{repl. corr.} & \textbf{no lex.} \\
\hline\hline
         Manual & 68.60         & 73.20 & 65.80         \\
%\hline 
\hline
         GPT-4o & 41.93          & 49.53  & 37.40        \\
%\hline
         GECToR & \textbf{43.73}          & \textbf{51.80}  & \textbf{42.53}       \\
%\hline
         Gramformer &  41.60        & 49.60  & 40.19       \\
\hline
        No GEC & 28.46         & 43.13   & 28.20       \\

\end{tabular}
\caption{Overall Accuracy (\%) results for \textbf{feedback-based} evaluation (standard, replace corrected word and no lexical information) when using GPT-4o both for GEF generation and discrimination.}
\label{tab:gef_res_feedback_gpt4}
\end{table}
In this case, we can see a large drop in performance for the No GEC system when compared to the others, while GECToR is still our best GEC system overall, and Gramformer outperforms GPT-4o when using the \emph{replaced corrected word} and \emph{no lexical information} \( M^2 \) files. The presence of lexical information, especially when we replace the corrections with the originally incorrect words, boosts the performance of all systems. Note that these figures are not directly comparable to those in Table \ref{tab:gef_res_essay_gpt4} because they have different lineups, i.e., the former has essays, while the latter has GEF responses. Therefore, the fact that, when using the \emph{replaced corrected word} format with manual corrections, the feedback-based evaluation method (73.20\%) apparently performs better than the essay-type-based method (61.26\%) should not mislead the reader.

\begin{figure}[!htbp]
    \centering
    \includegraphics[width=1.0\linewidth]{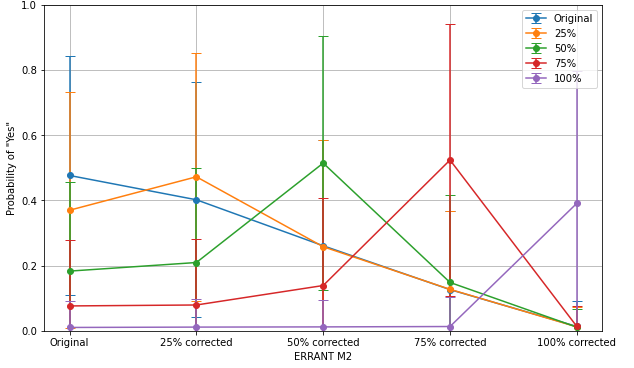}
    \caption{Mean probabilities of ``Yes'' for feedback-based evaluation (\emph{no lexical information}) - Manual GEC only.}
    \label{fig:mean_probs_man}
\end{figure}

More performance figures using the feedback-based evaluation method (\emph{no lexical information}) can be found in the confusion matrices in Figure \ref{fig:conf_matrices} (see Appendix \ref{sec:appendix_d}). These results are in line with the ones previously obtained on GEC: when using GPT-4o as GEC system, performances are slightly better on the \emph{original} and \emph{25\% corrected} essays, but they get worse as the number of errors decreases. Furthermore, an interesting perspective is offered by Figure \ref{fig:mean_probs_man}, which shows the mean probabilities of the first predicted token being ``Yes'' at varying error rates using manual corrections. It can be observed that the peaks are where we would expect them to be. Here, we only report these results using the manual corrections, but we can observe very similar figures for the other systems, as shown in Figure \ref{fig:probs_all} (see Appendix \ref{sec:appendix_d}). As can be expected, this trend is even more evident when we use the essay-type-based evaluation as well as when we use the \emph{standard} and \emph{replaced corrected word} formats of \( M^2 \) files, as these contain lexical information. However, due to space constraints, we cannot report these figures.

For the same reason, hereafter, we will focus on the analysis of the remaining results by only employing the feedback-based \emph{no lexical information} \( M^2 \)  format and essay-type-based evaluation methods, so to have a sort of a lower and an upper bound, respectively. As mentioned in Section \ref{sec: experimental}, we did not use only GPT-4o for GEF generation but also GPT-3.5 and Llama 3.

\begin{table}[h!]
\small
\centering
\begin{tabular}{c||ccc}
\multirow{3}{*}{} & \multicolumn{3}{c}{\textbf{\% Accuracy}} \\
%\cline{2-4}
\textbf{Form of GEC}            & \textbf{Llama 3} & \textbf{GPT-3.5} & \textbf{GPT-4o} \\
                    \hline\hline

Manual               & 43.40 &  52.26          & 61.26  \\ %\hline \hline

 GECToR & 42.73 &          49.86 & 59.93   \\ %\hline

 No GEC            & 37.86 &           42.26 & 57.46   \\

\end{tabular}
\caption{Overall Accuracy (\%) results comparing GPT-4o, GPT-3.5, and Llama 3 for GEF generation (essay-type-based).}
\label{tab:gef_gen_three_essay}
\end{table}

\begin{table}[h!]

\centering
\small
\begin{tabular}{c||ccc}
\multirow{3}{*}{} & \multicolumn{3}{c}{\textbf{\% Accuracy}} \\
% & \multicolumn{3}{c}{\textbf{no lex. information}} \\
%\cline{2-4}
 \textbf{Form of GEC}                  & \textbf{Llama 3} & \textbf{GPT-3.5} & \textbf{GPT-4o}  \\
\hline\hline
 Manual            & 36.00 & 47.93 & 65.80    \\ 

 GECToR            & 29.86 & 35.00 & 42.53    \\ %\hline

No GEC            & 25.00 & 27.20 & 28.20     \\

\end{tabular}
\caption{Overall Accuracy (\%) results comparing GPT-4o, GPT-3.5, and Llama 3 for GEF generation (feedback-based with no lexical information).}
\label{tab:gef_gen_three_feedback}
\end{table}

Tables \ref{tab:gef_gen_three_essay} and \ref{tab:gef_gen_three_feedback} compare the performances of the three LLMs when used for GEF generation employing the essay-type-based and the feedback-based (\emph{no lexical information}) evaluation methods, respectively. As can be observed, when we use Llama 3 and GPT-3.5, the results are significantly lower but very consistent with the results obtained using GPT-4o for GEF generation.

In order to rule out self-bias in the last part of our pipeline, i.e., GEF discrimination, we also try replacing GPT-4o with GPT-3.5 and Llama 3 in this final step.

\begin{table}[h!]
\small
\centering
\begin{tabular}{c||ccc}

 & \multicolumn{3}{c}{\textbf{\% Accuracy}} \\ 
%  & \multicolumn{3}{c}{\textbf{essay-type-based}} \\

%\cline{2-5}
\textbf{Form of GEF} & \textbf{Llama 3} & \textbf{GPT-3.5} & \textbf{GPT-4o} \\
\hline\hline
 Llama 3   & 23.39  & 23.66  & 43.40  \\ %\cline{2-5}
   GPT-3.5  & 28.80  & 25.53  & 52.26  \\ %\cline{2-5}
 GPT-4o   & 26.73  & 29.06  & 61.26  \\ 
\end{tabular}
\caption{Overall Accuracy (\%) results comparing GPT-4o, GPT-3.5, and Llama 3 for GEF discrimination with essay-type-based evaluation (Manual only).}
\label{tab:gec_discr_essay}
\end{table}

Tables \ref{tab:gec_discr_essay} and \ref{tab:gec_discr_feedback} show the results for both evaluation methods using the three LLMs. The performances are consistent with what we have observed thus far. The only exception arises when we use Llama 3 for GEF discrimination as it seems to prefer the GEF responses generated by GPT-3.5 instead of itself or GPT-4o. However, this does not compromise the consistency of the results because we still do not observe any type of self-bias.

\begin{table}[h!]
\small
\centering
\begin{tabular}{c||ccc}

 & \multicolumn{3}{c}{\textbf{\% Accuracy}} \\ 
%  & \multicolumn{3}{c}{\textbf{essay-type-based}} \\

%\cline{2-5}
\textbf{Form of GEF} & \textbf{Llama 3} & \textbf{GPT-3.5} & \textbf{GPT-4o} \\ \hline\hline
 Llama 3  & 17.06   & 22.00   & 36.00   \\ %\cline{2-5}
GPT-3.5   & 20.99   & 25.20   & 47.93   \\ %\cline{2-5}
GPT-4o  & 20.26  & 26.66   & 65.80  \\ 
\end{tabular}
\caption{Overall Accuracy (\%) results comparing GPT-4o, GPT-3.5, and Llama 3 for GEF discrimination with feedback-based evaluation (Manual only).}
\label{tab:gec_discr_feedback}
\end{table}

Finally, we test the sensitivity of the system to the lineup, by increasing its foils. For the sake of brevity and clarity, we only show these results when using the manual corrections and GPT-4o for GEF generation and discrimination and employing the feedback-based evaluation method with \( M^2 \) files without lexical information. In addition to \emph{original}, \emph{25\% corrected}, \emph{50\% corrected}, \emph{75\% corrected}, and \emph{100\% corrected}, we also consider four additional essay versions setting the error rate at 15\%, 40\%, 60\%, and 85\%. When increasing the number of elements in the lineup to 9, as expected, the overall Accuracy decreases from 65.80\% to 54.00\%, but the confusion matrix reported in Figure \ref{fig:conf_matrix_9_foils} (see Appendix \ref{sec:appendix_d}) shows a leading diagonal that is consistent with the one related to the previous lineup of 5 elements (see Figure \ref{fig:conf_matrices}a for comparison). Additionally, in Figure \ref{fig:mean_probs_9_foils} (see Appendix \ref{sec:appendix_d}), we also report the mean probabilities of the first predicted token being ``Yes'' at varying error rates after increasing the lineup elements to 9, similarly to what we did in Figure \ref{fig:mean_probs_man}. We can still observe very consistent results.

\section{Discussion}
\label{sec:discussion}

In this study, we found that providing GEC information to the LLM is fundamental in order for it to produce correct feedback. Using a sort of a chain-of-thought approach, whereby we also use the GEC version of an essay, helps the LLM provide a better GEF response. On the other hand, when GEF generation cannot exploit the information derived from GEC, we generally observe significantly lower results. Furthermore, having a high-performing GEC system improves the quality of a GEF response. In our experiments, we found that using GECToR achieves better results than using GPT-4o. These results also exclude the presence of self-bias derived from the GEC step. Self-bias is also ruled out in the other two steps of the pipeline since we replaced GPT-4o with GPT-3.5 and Llama 3 and still obtained consistent results. Additionally, in our study we showed that our implicit evaluation approach is particularly effective for GEF because it is intrinsically: a) cheap because it does not require manual GEF annotations; b) flexible because foils and lineups can be customised.

Finally, when it comes to choosing one of the evaluation methods proposed in the study, we need to keep in mind that LLMs seem to be strongly biased by lexical information. Therefore, we used a feedback-based evaluation method with edited \( M^2 \) files in which the system does not exploit lexical information derived from elements in the lineup.

\section{Conclusions and future work}
\label{sec:conclusions}

In this paper, we presented a novel implicit evaluation framework to provide grammatical error feedback to L2 learner essays in a compact, comprehensive, and unstructured form. After editing our dataset by varying the error rates of its essays, our proposed pipeline is divided into three major steps: GEC, GEF generation, and GEF discrimination. This framework is cheaper than those previously proposed as it does not require human feedback annotations but only parallel original and corrected essays. Furthermore, it is quite flexible since it is based on lineups of various essay versions and feedback responses derived from these, which can be modified and customised.

Future work will extend this framework to written data in other languages and spoken data, possibly by including multimodal LLMs, such as SALMONN~\citep{tang2024salmonn} and Qwen-Audio \citep{Qwen-Audio}.

\section*{Acknowledgments}

This paper reports on research supported by Cambridge University Press \& Assessment, a department of The Chancellor, Masters, and Scholars of the University of Cambridge. The authors would like to thank the ALTA Spoken Language Processing Technology Project Team for general discussions and contributions to the evaluation infrastructure.

\section*{Limitations}

Arguably, the main limitation of the experiments outlined in this paper is the lack of a reference set of feedback responses in absolute terms since our evaluation approach is implicit by definition. Also, the need for GEC annotations might constitute an issue at first glance, but it seems to be a valid trade-off compared to needing GEF annotations, which are significantly more costly.

\section*{Ethical considerations}

We confirm that this work adheres to the ACL Code of Ethics. We acknowledge the potential biases inherent in pre-trained LLMs and took steps to mitigate these by evaluating our models across various learner L1s, nationalities, and proficiency levels. This research was conducted with the goal of enhancing educational tools and supporting language learners, and we remain committed to continuous ethical reflection and improvement in our methods and applications.

% Bibliography entries for the entire Anthology, followed by custom entries
%\bibliography{anthology,custom}
% Custom bibliography entries only

\bibliography{custom}

\appendix

\section{Appendix A: Prompts}
\label{sec:appendix_a}

\subsection{Prompt for GEC}

The prompt given to GPT-4o for GEC is the following:

\begin{quote}
    \texttt{Read the following essay written by an L2 learner of English: [ESSAY]}
    
    \texttt{Provide the grammatically corrected version of the essay without adding any comment, note, or explanation.}
\end{quote}

\subsection{Prompts for GEF generation}

\paragraph{With GEC:} the prompt used for GEF generation when exploiting the information derived from GEC is the following:

\begin{quote}
    \texttt{Read the following essay written by an L2 learner of English and its respective corrected version:}
    
    \texttt{Original: [ESSAY]}
    
    \texttt{Corrected: [CORRECTED ESSAY]}
    
    \texttt{Provide grammatical feedback to the learner based on the differences between the original and the corrected version. Start your feedback with ``Dear learner''.}
\end{quote}

When using this prompt on the \emph{100\% corrected} version of the essays, we noticed that the systems tended to also include a revised version of the essays along with the feedback response. Therefore, in such cases, we slightly modified the prompt as follows:

\begin{quote}
    \texttt{Read the following essay written by an L2 learner of English and its respective corrected version:}
    
    \texttt{Original: [ESSAY]}
    
    \texttt{Corrected: [CORRECTED ESSAY]}
    
    \texttt{Provide grammatical feedback to the learner based on the differences between the original and the corrected version. You don't have to provide a revised version of the essay. Start your feedback with ``Dear learner''.}
\end{quote}

\paragraph{Without GEC:} the prompt used for GEF generation without including the information derived from GEC is the following:

\begin{quote}
    \texttt{Read the following essay written by an L2 learner of English: [ESSAY]}
    
    \texttt{Provide grammatical feedback to the learner. You don't have to provide a revised version of the essay. Start your feedback with ``Dear learner''.}
\end{quote}

\subsection{Prompts for GEF discrimination}

\paragraph{Prompt for essay-type-based evaluation:}

\begin{quote}
    \texttt{Read the following essay written by an L2 learner of English: [ESSAY]}
    
    \texttt{Now, read the following feedback: [FEEDBACK]}
    
    \texttt{Is it correct, appropriate and thorough? Just say yes or no without providing any additional comment, note or explanation.}

\end{quote}

\paragraph{Prompt for feedback-based evaluation:}

\begin{quote}
    \texttt{Read the following feedback response to an essay written by an L2 learner: [FEEDBACK]}
    
    \texttt{Now, read this ERRANT (ERRor ANnotation Toolkit) file: [ERRANT M2 FILE]}
    
    \texttt{in which M: missing, R: replace, U: unnecessary, ADJ: adjective, ADV: adverb, CONJ: conjunction, DET: determiner, NOUN: noun, PART: particle, PREP: preposition, PRON: pronoun, PUNCT: punctuation, VERB: verb, CONTR: contraction, OTHER: other, MORPH: morphology, ORTH: orthography, SPELL: spelling, WO: word order, FORM: form, INFL: inflection, NUM: number, POSS: possessive, SVA: agreement, TENSE: tense.}
    
    \texttt{Does the feedback response explain the ERRANT file correctly, appropriately and thoroughly? Just say yes or no without providing any additional comment, note or explanation.}

\end{quote}

\section{Appendix B: GEF example}
\label{sec:appendix_b}

The following is an example of GEF response obtained using GPT-4o:

\begin{quote}
    \texttt{Dear learner,}

\texttt{Thank you for sharing your essay. Here are some grammatical points to help you improve your writing based on the differences between your original version and the corrected version:}

\texttt{1. **Spelling**: 
   - In your original sentence, you wrote ``liste'' instead of ``listen.'' Make sure to double-check the spelling of words to ensure accuracy.}

\texttt{2. **Verb Choice**:
   - You used ``see'' in ``I can see funny videos.'' While ``see'' is not incorrect, ``watch'' is a more appropriate verb when referring to viewing videos or TV shows. ``Watch'' implies that you are actively paying attention to the videos.}

\texttt{3. **Verb Tense**:
   - You wrote ``It is expensive'' in your original sentence. The corrected version uses ``It was expensive.'' This change is made because you are referring to the cost of the phone at the time of purchase, which is in the past. Using the past tense ``was'' is more accurate in this context.}

\texttt{By paying attention to these details, you can make your writing clearer and more precise. Keep practicing, and you will continue to improve!}

\texttt{Best regards,
[Your Name]}
\end{quote}

This was generated after feeding the \emph{original} and manually corrected versions of this essay written by an A1 learner:\footnote{Note that we chose to report a short essay due to space constraints.}

\begin{quote}
    \textbf{Original:} \emph{Hello Mike,}
    
    \emph{I bought a mobile phone, I like it because I can liste to music and I can see funny videos. It is expensive but it is good and it is black.}
    
    \emph{Write soon.}

    \textbf{Correct:} \emph{Hello Mike,}
    
    \emph{I bought a mobile phone, I like it because I can listen to music and I can watch funny videos. It was expensive but it is good and it is black.}
    
    \emph{Write soon.}
\end{quote}

\section{Appendix C: ERRANT M2 files}
\label{sec:appendix_c}

The following examples are related to the essay reported in Appendix \ref{sec:appendix_b} and can be matched with the example of the original ERRANT M2 file shown in Section \ref{sec:lineup} in order to have a comprehensive view.

\subsection{Standard}

\texttt{A 15 16|||R:SPELL|||listen|||REQUIRED|||-NONE-|||0
A 21 22|||R:VERB|||watch|||REQUIRED|||-NONE-|||0
A 26 27|||R:VERB:TENSE|||was|||REQUIRED|||-NONE-|||0}

\subsection{Replaced corrected word}

\texttt{A 15 16|||R:SPELL|||liste|||REQUIRED|||-NONE-|||0
A 21 22|||R:VERB|||see|||REQUIRED|||-NONE-|||0
A 26 27|||R:VERB:TENSE|||is|||REQUIRED|||-NONE-|||0}

\subsection{No lexical information}

\texttt{A 15 16|||R:SPELL||||||REQUIRED|||-NONE-|||0
A 21 22|||R:VERB||||||REQUIRED|||-NONE-|||0
A 26 27|||R:VERB:TENSE||||||REQUIRED|||-NONE-|||0}

\section{Appendix D: Further experimental results}
\label{sec:appendix_d}

In this section, we report further experimental results. Specifically, Tables \ref{tab:gec_res_full_manual}, \ref{tab:gec_res_full_gpt4o}, \ref{tab:gec_res_full_gector}, and \ref{tab:gec_res_full_gram} show the results for GEC in terms of Precision, Recall, and F$_{0.5}$ scores for the Manual, GPT-4o, GECToR, and Gramformer GEC systems, respectively.

\begin{figure*}[htbp]
    \centering
    \subfloat[Manual GEC]{
        \includegraphics[width=0.5\textwidth]{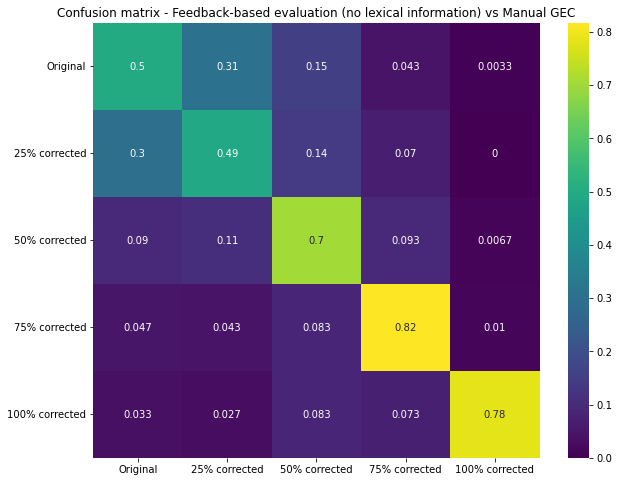}
        \label{subfig:figure1}
    }
    \subfloat[Auto GEC (GPT-4o)]{
        \includegraphics[width=0.5\textwidth]{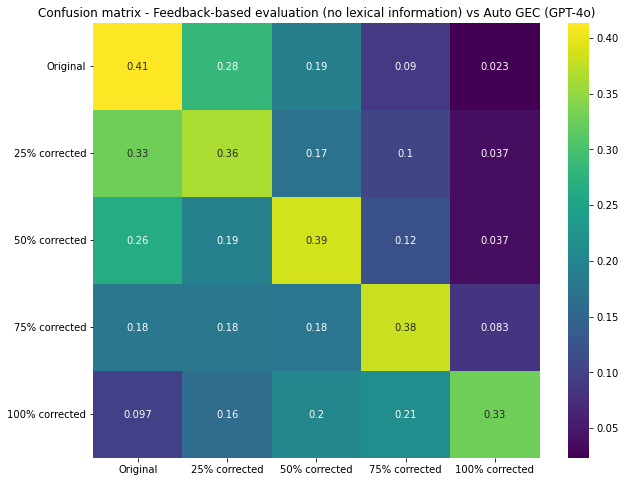}
        \label{subfig:figure2}
    }\\
    \subfloat[Auto GEC (GECToR)]{
        \includegraphics[width=0.5\textwidth]{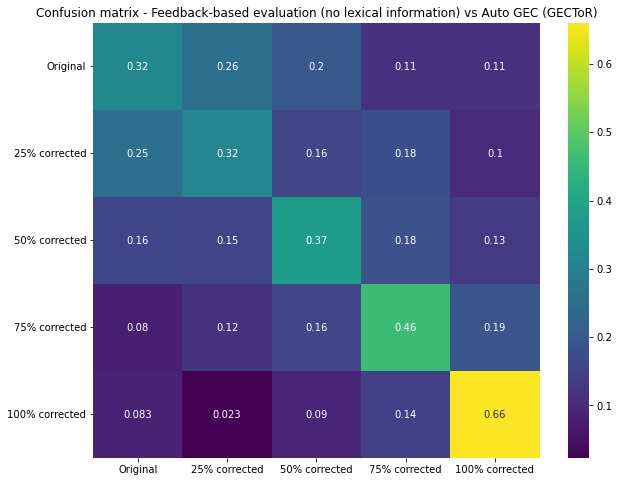}
        \label{subfig:figure3}
    }
    \subfloat[Auto GEC (Gramformer)]{
        \includegraphics[width=0.5\textwidth]{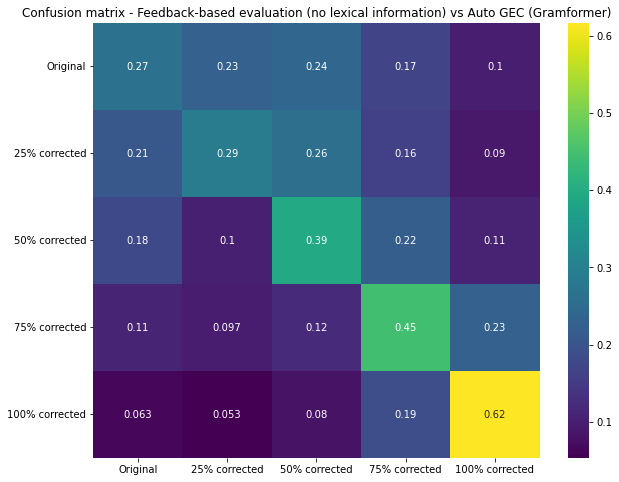}
        \label{subfig:figure4}
    }\\
    \subfloat[No GEC]{
        \includegraphics[width=0.5\textwidth]{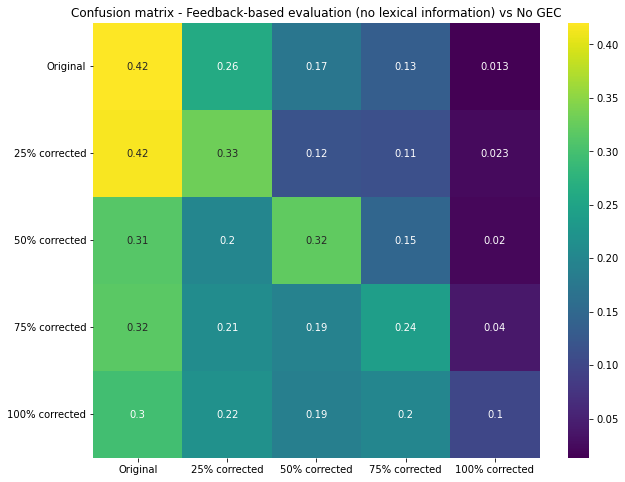}
        \label{subfig:figure5}
    }
    \caption{Confusion matrices for five GEC systems with feedback-based evaluation method (\emph{no lexical information}): manual GEC, GPT-4o, GECToR, Gramformer, and No GEC.}
    \label{fig:conf_matrices}
\end{figure*}

\begin{figure*}[!htbp]
    \centering
    \includegraphics[width=1.0\linewidth]{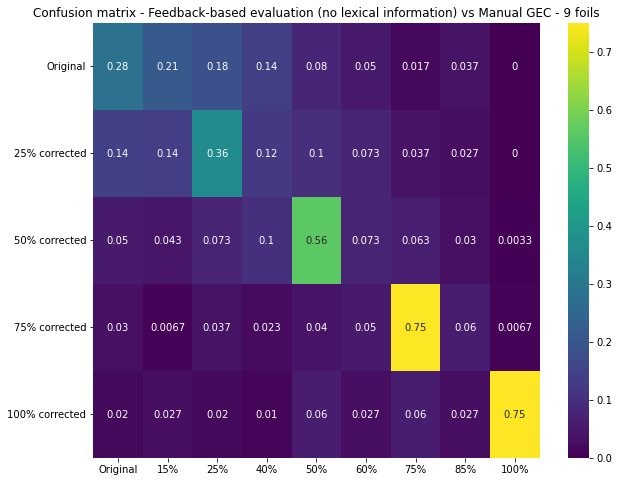}
    \caption{Confusion matrix - Feedback-based evaluation (\emph{no lexical information}) - Manual GEC only.}
    \label{fig:conf_matrix_9_foils}
\end{figure*}

Figure \ref{fig:conf_matrices} shows the confusion matrices for five GEC systems (i.e., Manual, GPT-4o, GECToR, Gramformer, and No GEC) with feedback-based evaluation method (\emph{no lexical information}). Similarly, Figure \ref{fig:conf_matrix_9_foils} shows the confusion matrix for the Manual GEC system with feedback-based evaluation method (\emph{no lexical information}) using a lineup of 9 elements.

\begin{figure*}[htbp]
    \centering
    \subfloat[Feedback original]{
        \includegraphics[width=0.5\textwidth]{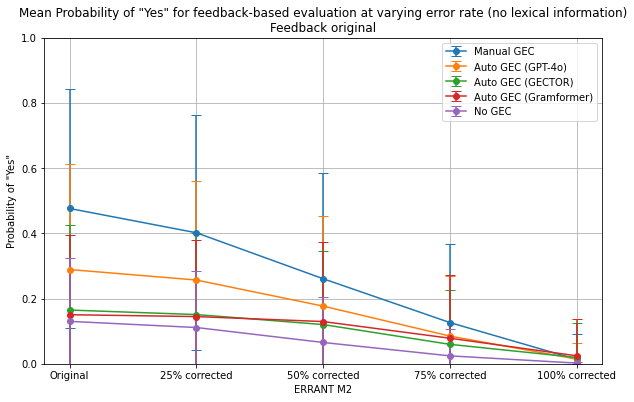}
        \label{subfig:figure1mean}
    }
    \subfloat[Feedback 25\% corrected]{
        \includegraphics[width=0.5\textwidth]{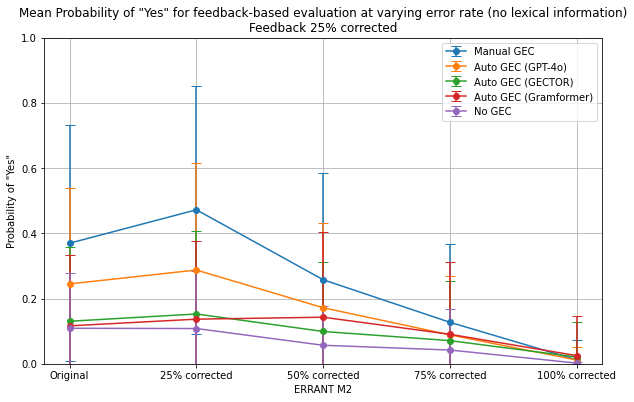}
        \label{subfig:figure2mean}
    }\\
    \subfloat[Feedback 50\% corrected]{
        \includegraphics[width=0.5\textwidth]{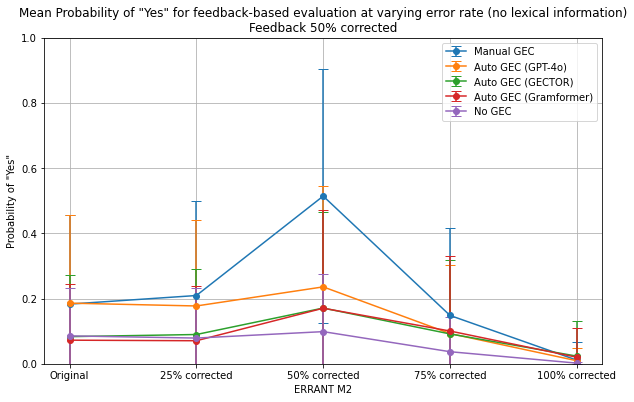}
        \label{subfig:figure3mean}
    }
    \subfloat[Feedback 75\% corrected]{
        \includegraphics[width=0.5\textwidth]{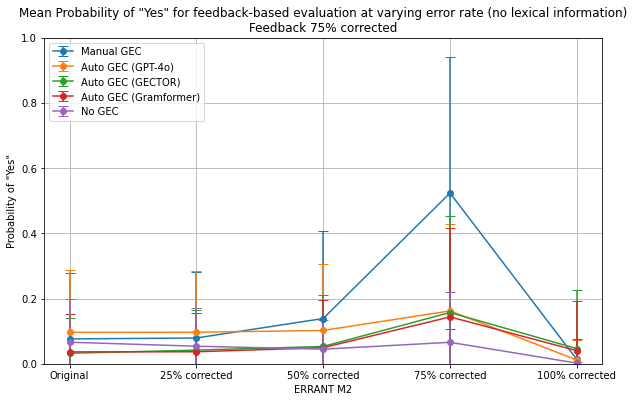}
        \label{subfig:figure4mean}
    }\\
    \subfloat[Feedback 100\% corrected]{
        \includegraphics[width=0.5\textwidth]{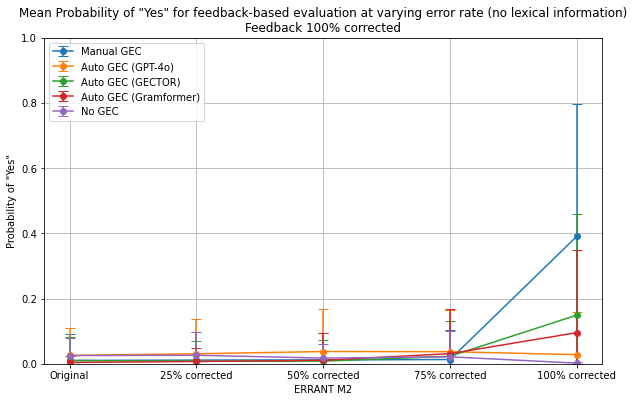}
        \label{subfig:figure5mean}
    }
    \caption{Mean probabilities of ``Yes'' for feedback-based evaluation at varying error rate (no lexical information).}
    \label{fig:probs_all}
\end{figure*}

\begin{figure*}[!htbp]
    \centering
    \includegraphics[width=1.0\linewidth]{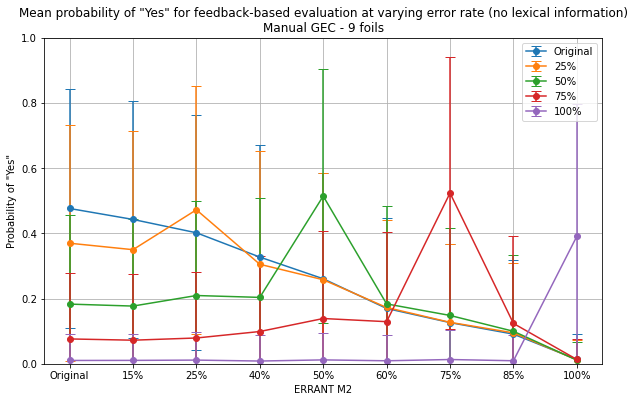}
    \caption{Mean probabilities of “Yes” for feedback-based evaluation (no lexical information) - Manual GEC
only - 9 elements.}
    \label{fig:mean_probs_9_foils}
\end{figure*}

Figure \ref{fig:probs_all} illustrates the mean probabilities for ``Yes'' being the first predicted token by GPT-4o in the GEF discrimination step at varying error rate for all the GEC systems considered in our experiments. In this case, we also used the feedback-based evaluation method (\emph{no lexical information}). Figure \ref{fig:mean_probs_9_foils} shows the mean probabilities of ``Yes'' being the first predicted token for the Manual GEC system only when using the feedback-based evaluation (\emph{no lexical information}) and 9 elements in the lineup.

%PROBS

\clearpage

\begin{table*}[h!]
\centering

\footnotesize

  \begin{tabular}{c|c|c|c|c}
    \hline
    \multirow{2}{*}{\textbf{HYP/REF}} & \multicolumn{4}{c}{\textbf{Manual}} \\ \cline{2-5}
                                      & \textbf{Original} & \textbf{25\%} & \textbf{50\%} & \textbf{75\%}  \\ \hline
    \textbf{Original} & \textcolor{red}{100.00}/\textcolor{orange}{100.00}/\textcolor{blue}{100.00} & \textcolor{red}{37.11}/\textcolor{orange}{47.21}/\textcolor{blue}{38.77} & \textcolor{red}{18.20}/\textcolor{orange}{34.31}/\textcolor{blue}{20.09} & \textcolor{red}{9.29}/\textcolor{orange}{31.96}/\textcolor{blue}{10.82}  \\ \hline
    \textbf{25\%} & \textcolor{red}{47.21}/\textcolor{orange}{37.11}/\textcolor{blue}{44.77} & \textcolor{red}{100.00}/\textcolor{orange}{100.00}/\textcolor{blue}{100.00} & \textcolor{red}{19.05}/\textcolor{orange}{28.22}/\textcolor{blue}{20.37} & \textcolor{red}{9.94}/\textcolor{orange}{26.87}/\textcolor{blue}{11.37} \\ \hline
    \textbf{50\%} & \textcolor{red}{34.31}/\textcolor{orange}{18.20}/\textcolor{blue}{29.15} & \textcolor{red}{28.22}/\textcolor{orange}{19.05}/\textcolor{blue}{25.74} & \textcolor{red}{100.00}/\textcolor{orange}{100.00}/\textcolor{blue}{100.00} & \textcolor{red}{12.41}/\textcolor{orange}{22.65}/\textcolor{blue}{13.65} \\ \hline
    \textbf{75\%} & \textcolor{red}{31.96}/\textcolor{orange}{9.29}/\textcolor{blue}{21.47} & \textcolor{red}{26.87}/\textcolor{orange}{9.94}/\textcolor{blue}{20.04} & \textcolor{red}{22.65}/\textcolor{orange}{12.41}/\textcolor{blue}{19.44} & \textcolor{red}{100.00}/\textcolor{orange}{100.00}/\textcolor{blue}{100.00}  \\ 
  \end{tabular}
  \caption{ERRANT \textcolor{red}{Precision}, \textcolor{orange}{Recall}, and  \textcolor{blue}{F$_{0.5}$} scores for \textbf{Manual} vs Manual.}

\label{tab:gec_res_full_manual}

\medskip
  
    \begin{tabular}{c|c|c|c|c}
    \hline
    \multirow{2}{*}{\textbf{HYP/REF}} & \multicolumn{4}{c}{\textbf{Manual}} \\ \cline{2-5}
                                      & \textbf{Original} & \textbf{25\%} & \textbf{50\%} & \textbf{75\%} \\ \hline
    \textbf{Original} & \textcolor{red}{41.77}/\textcolor{orange}{52.45}/\textcolor{blue}{43.55} & \textcolor{red}{16.65}/\textcolor{orange}{26.60}/\textcolor{blue}{18.00} & \textcolor{red}{7.58}/\textcolor{orange}{19.73}/\textcolor{blue}{8.57} & \textcolor{red}{3.97}/\textcolor{orange}{17.14}/\textcolor{blue}{4.69} \\ \hline
    \textbf{25\%} & \textcolor{red}{18.43}/\textcolor{orange}{20.78}/\textcolor{blue}{18.86} & \textcolor{red}{37.84}/\textcolor{orange}{54.28}/\textcolor{blue}{40.28} & \textcolor{red}{7.53}/\textcolor{orange}{16.00}/\textcolor{blue}{8.42} & \textcolor{red}{3.91}/\textcolor{orange}{15.16}/\textcolor{blue}{4.59} \\ \hline
    \textbf{50\%} & \textcolor{red}{10.27}/\textcolor{orange}{9.99}/\textcolor{blue}{10.22} & \textcolor{red}{8.96}/\textcolor{orange}{11.09}/\textcolor{blue}{9.32} & \textcolor{red}{30.61}/\textcolor{orange}{56.11}/\textcolor{blue}{33.67} & \textcolor{red}{4.12}/\textcolor{orange}{13.78}/\textcolor{blue}{4.79}  \\ \hline
    \textbf{75\%} & \textcolor{red}{6.48}/\textcolor{orange}{5.28}/\textcolor{blue}{6.20} & \textcolor{red}{5.47}/\textcolor{orange}{5.67}/\textcolor{blue}{5.51} & \textcolor{red}{4.82}/\textcolor{orange}{7.41}/\textcolor{blue}{5.19} & \textcolor{red}{20.16}/\textcolor{orange}{56.50}/\textcolor{blue}{23.14}  \\
    
  \end{tabular}
  \caption{ERRANT \textcolor{red}{Precision}, \textcolor{orange}{Recall}, and  \textcolor{blue}{F$_{0.5}$} scores for \textbf{Auto GEC (GPT-4o)} vs Manual.}
\label{tab:gec_res_full_gpt4o}

\medskip

\begin{tabular}{c|c|c|c|c}
    \hline
    \multirow{2}{*}{\textbf{HYP/REF}} & \multicolumn{4}{c}{\textbf{Manual}} \\ \cline{2-5}
                                      & \textbf{Original} & \textbf{25\%} & \textbf{50\%} & \textbf{75\%}\\ \hline
    \textbf{Original}                 & \textcolor{red}{44.10}/\textcolor{orange}{31.17}/\textcolor{blue}{40.90} & \textcolor{red}{17.83}/\textcolor{orange}{16.31}/\textcolor{blue}{17.51} & \textcolor{red}{7.73}/\textcolor{orange}{10.48}/\textcolor{blue}{8.16} & \textcolor{red}{4.04}/\textcolor{orange}{9.99}/\textcolor{blue}{4.59} \\ \hline
    \textbf{25\%}                     & \textcolor{red}{20.12}/\textcolor{orange}{12.74}/\textcolor{blue}{18.03} & \textcolor{red}{41.11}/\textcolor{orange}{33.13}/\textcolor{blue}{39.22} & \textcolor{red}{7.31}/\textcolor{orange}{8.73}/\textcolor{blue}{7.56} & \textcolor{red}{4.39}/\textcolor{orange}{9.56}/\textcolor{blue}{4.92}  \\ \hline
    \textbf{50\%}                     & \textcolor{red}{11.33}/\textcolor{orange}{5.86}/\textcolor{blue}{9.54} & \textcolor{red}{9.68}/\textcolor{orange}{6.37}/\textcolor{blue}{8.77} & \textcolor{red}{34.08}/\textcolor{orange}{33.22}/\textcolor{blue}{33.90} & \textcolor{red}{4.26}/\textcolor{orange}{7.58}/\textcolor{blue}{4.67} \\ \hline
    \textbf{75\%}                     & \textcolor{red}{7.41}/\textcolor{orange}{3.10}/\textcolor{blue}{5.80} & \textcolor{red}{6.58}/\textcolor{orange}{3.50}/\textcolor{blue}{5.59} & \textcolor{red}{5.02}/\textcolor{orange}{3.96}/\textcolor{blue}{4.77} & \textcolor{red}{24.51}/\textcolor{orange}{35.31}/\textcolor{blue}{26.10} \\
    
  \end{tabular}
  
  \caption{ERRANT \textcolor{red}{Precision}, \textcolor{orange}{Recall}, and  \textcolor{blue}{F$_{0.5}$} scores for \textbf{Auto GEC (GECToR)} vs Manual.}
  
\label{tab:gec_res_full_gector}

\medskip

\begin{tabular}{c|c|c|c|c}
    \hline
    \multirow{2}{*}{\textbf{HYP/REF}} & \multicolumn{4}{c}{\textbf{Manual}} \\ \cline{2-5}
                                      & \textbf{Original} & \textbf{25\%} & \textbf{50\%} & \textbf{75\%}\\ \hline
    \textbf{Original}                 & \textcolor{red}{40.47}/\textcolor{orange}{24.39}/\textcolor{blue}{35.75} & \textcolor{red}{16.33}/\textcolor{orange}{12.52}/\textcolor{blue}{15.39} & \textcolor{red}{7.81}/\textcolor{orange}{8.87}/\textcolor{blue}{8.00} & \textcolor{red}{4.15}/\textcolor{orange}{8.61}/\textcolor{blue}{4.63} \\ \hline
    \textbf{25\%}                     & \textcolor{red}{18.45}/\textcolor{orange}{10.09}/\textcolor{blue}{15.83} & \textcolor{red}{37.27}/\textcolor{orange}{25.93}/\textcolor{blue}{34.27} & \textcolor{red}{7.14}/\textcolor{orange}{7.36}/\textcolor{blue}{7.19} & \textcolor{red}{4.08}/\textcolor{orange}{7.67}/\textcolor{blue}{4.50}  \\ \hline
    \textbf{50\%}                     & \textcolor{red}{10.86}/\textcolor{orange}{5.11}/\textcolor{blue}{8.86} & \textcolor{red}{9.32}/\textcolor{orange}{5.58}/\textcolor{blue}{8.22} & \textcolor{red}{32.11}/\textcolor{orange}{28.46}/\textcolor{blue}{31.31} & \textcolor{red}{3.99}/\textcolor{orange}{6.46}/\textcolor{blue}{4.32} \\ \hline
    \textbf{75\%}                     & \textcolor{red}{7.37}/\textcolor{orange}{2.75}/\textcolor{blue}{5.52} & \textcolor{red}{6.64}/\textcolor{orange}{3.15}/\textcolor{blue}{5.44} & \textcolor{red}{5.16}/\textcolor{orange}{3.63}/\textcolor{blue}{4.76} & \textcolor{red}{21.65}/\textcolor{orange}{27.82}/\textcolor{blue}{22.65} \\
    
  \end{tabular}
  
  \caption{ERRANT \textcolor{red}{Precision}, \textcolor{orange}{Recall}, and  \textcolor{blue}{F$_{0.5}$} scores for \textbf{Auto GEC (Gramformer)} vs Manual.}
\label{tab:gec_res_full_gram}
  
\end{table*}

\end{document}